\documentclass[sigconf]{acmart}
\usepackage{subfigure}
\usepackage{lipsum} 
\usepackage{algorithm}
\usepackage{algpseudocode}
\usepackage{balance}

\AtBeginDocument{%
  \providecommand\BibTeX{{%
    \normalfont B\kern-0.5em{\scshape i\kern-0.25em b}\kern-0.8em\TeX}}}

\begin{document}

\title{Typing on Any Surface: A Deep Learning-based Method for Real-Time Keystroke Detection in Augmented Reality}

\author{Xingyu Fu}
\email{xingyu.fu@data61.csiro.au}
\orcid{0000-0002-2535-7757}
\affiliation{%
  \institution{Data61, CSIRO}
  \institution{Australian National University}
  \city{Canberra}
  \state{ACT}
  \country{Australia}
  \postcode{2601}
}

\author{Mingze Xi}
\authornote{Corresponding author.}
\email{mingze.xi@data61.csiro.au}
\orcid{0000-0003-1291-4136}
\affiliation{%
  \institution{Data61, CSIRO}
  \institution{Australian National University}
  \city{Canberra}
  \state{ACT}
  \country{Australia}
  \postcode{2601}
}


\begin{abstract}
Frustrating text entry interface has been a major obstacle in participating in social activities in augmented reality (AR). Popular options, such as mid-air keyboard interface, wireless  keyboards or voice input, either suffer from poor ergonomic design, limited accuracy, or are simply embarrassing to use in public. This paper proposes and validates a deep-learning based approach, that enables AR applications to accurately predict keystrokes from the user perspective RGB video stream that can be captured by any AR headset. This enables a user to perform typing activities on any flat surface and eliminates the need of a physical or virtual keyboard. A two-stage model, combing an off-the-shelf hand landmark extractor and a novel adaptive Convolutional Recurrent Neural Network (C-RNN), was trained using our newly built dataset. The final model was capable of adaptive processing user-perspective video streams at ~32 FPS. This base model achieved an overall accuracy of $91.05\%$ when typing 40 Words per Minute (wpm), which is how fast an average person types with two hands on a physical keyboard. The Normalised Levenshtein Distance also further confirmed the real-world applicability of that our approach. The promising results highlight the viability of our approach and the potential for our method to be integrated into various applications. We also discussed the limitations and future research required to bring such technique into a production system.
\end{abstract}



\maketitle

\section{Introduction}
With new headsets being released at an increasing frequency, Augmented Reality (AR) is becoming more accessible to the general public. AR headsets, regardless it is optical see-through (e.g., Microsoft HoloLens 2 and Magic Leap 2) or video pass-through (e.g., Meta Quest Pro and Apple Vision Pro), allow users to interact with virtual content in their physical environment. Although big tech companies are heavily promoting the user engagement in social and professional activities in AR and Virtual Reality (VR), there are still technically challenging problems that need to be solved. One of them is the lack of a suitable text entry method.

There are several categories of text-input interfaces. Voice-based text entry methods are universally supported in modern devices. With advanced deep learning, this method has been improved significantly in recent years. There are a number of popular models or APIs that can be directly integrated into AR systems, such as Whisper~\cite{whisperOpenAI} from OpenAI, Microsoft Azure Speech Services and Google Cloud Speech recognition. The biggest issue with these speech-to-text (STT) methods is the privacy concern when used in a public space. Some users even feel embarrassed to use them in their private environment, let along in the public. Meanwhile, it is still less satisfying for noisy environments or for inputting non-vocabulary items, and it also poses latency issues.

Mid-air keyboard is a staple way of entering text in AR HMDs, such as the system keyboard in HoloLens 2 and Magic Leap 2. This involves positioning a virtual keyboard in front of a user, and the user can type by tapping on the virtual button. Due to technical limitations, early devices also used gaze~\cite{Xu2019} or third-party eye-trackers~\cite{Mott2017} as a pointer/cursor, instead of tapping directly on the virtual keyboard. However, this method is not suitable for long-term use due to the lack of tactile feedback and the fatigue caused by the unnatural hand/arm posture.

Some research studies also used customised wearable devices to capture user hand motion data, which was used to predict user keystrokes, such as the wristbands used in TapType~\cite{Streli2022} and the ring device used in QwertyRing~\cite{Gu2020}.

Noticeably, there is a common problem that all of them share: the low typing speed. According to several online type speed testing websites, 40 words per minute (wpm) is considered to be the average typing speed for English speakers. However, the fastest typing speed achieved in AR is only around 30 wpm~\cite{Streli2022}. This is a major obstacle for AR users to participate in social activities, such as chatting, taking notes, and writing emails.

Despite these pioneering efforts, no ideal solution has been found that provides a natural and intuitive interface, with decent typing speed, tactile feedback, and better privacy. Thus, there exists a clear need for a new text input method that addresses these gaps. One way to approach this is to use neural networks to predict user keystrokes from hand movements. 

In this paper, we propose an early design of a neural network based approach that enables the AR headsets to accurately predict keystrokes from hand movements, while the user is typing on any flat surface without a physical or virtual keyboard. Instead of fine-tuning a production model for a specific headset, we experimented with a two-stage model, combing an off-the-shelf hand landmark extractor and an adaptive Convolutional Recurrent Neural Network (C-RNN). The hand landmark detection stage could be easily replaced with on-device hand tracking models in a production system. In addition, we also present a custom-built dataset, AR Keystroke Detection Dataset (AKDD), as there is no existing dataset that is suitable for our task.

The rest of the paper is organised as follows. Section~\ref{sec:related_works} looks into previous work on text entry in AR. Section~\ref{sec:data_collection} describes the data collection process and the collected dataset. Section~\ref{sec:model_design} presents our two-stage keystroke detection pipeline. Benchmarking results are shown and discussed in Section~\ref{sec:results}. Finally, we summarised the main findings and point out future directions in Section~\ref{sec:conclusions}.

\section{Related Works} \label{sec:related_works}
Unlike Virtual Reality Head-Mounted Displays (VR HMDs), where users cannot see the real world, AR HMDs overlay virtual content on the user's physical environment, allowing users to see their surroundings. Therefore, traditional text entry methods, such as a physical keyboard, can be directly connected to an AR headset. Using such devices minimises the learning curve and provides a familiar experience for users. However, an external device is not always available and requires additional purchases and setups, and it may also limit the user's mobility. This section revisits previous attempts to avoid the use of a physical keyboard by exploiting other channels.

\subsection{Wearables-based Text Entry} \label{subsec:wearables}
\begin{figure}[h]
  \centering
  \includegraphics[width=0.9\linewidth]{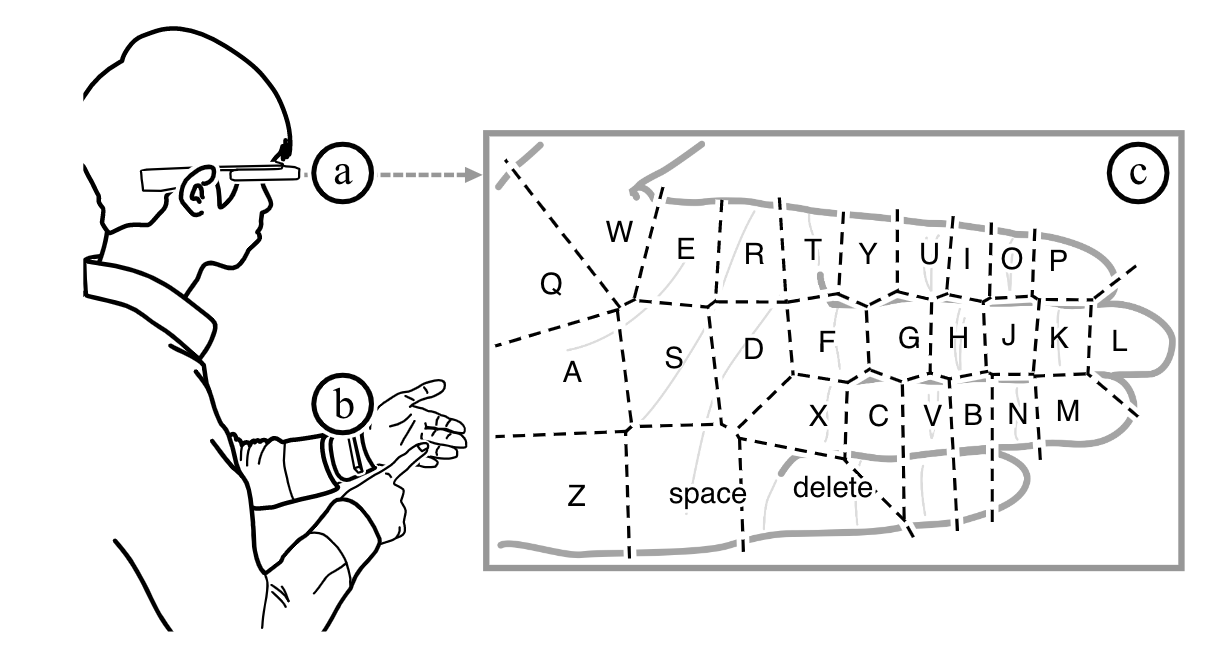}
  \caption{The schematic diagram of PalmType~\citep{Wang2015}.}
  \label{fig:PalmType}
\end{figure}

There have been a handful of studies on conducting text entry using wearable devices. An early example is the PalmType~\cite{Wang2015}, which used a wearable display that projects a layout-optimised virtual keyboard on the user's palm. A wrist-worn sensor detects the user's finger movements to input text (see Figure~\ref{fig:PalmType}). The user can type by tapping the virtual keys with their fingers. It achieved a typing speed of 7.7 wpm. DigiTap uses a fisheye camera and an accelerometer worn on the user's wrist to capture thumb-to-finger gestures and translate them into text. It achieved a typing speed of approximately 10 wpm~\cite{Pick2016}.

Apart from wristbands, Gu et al.~\cite{Gu2020} created QwertyRing using a customised ring device and a Bayesian decoder to predict which key the user was tapping. 
With sufficient practice, participants could achieve a typing speed of 20.59 wpm. Another recent work involves wristbands and machine learning is TapType~\cite{Streli2022}, which is a typing system that allows full-size typing on a flat surface. Two vibration sensors are placed on the user's wrists, and a Bayesian classifier is used to estimate the tapping finger. Then an n-gram language model is used to predict the character. It achieved an average typing speed of 19 wpm. Similarly, Kwon et al. used MYO armband and a neural network to predict the keyboard input~\cite{Kwon2020}. The system achieved a typing speed of 20.5 wpm.

A noticeable issue with these methods is the relatively low typing speeds compared to the traditional keyboard (approx. 40 wpm). Meanwhile, it also requires dedicated effort to learn how to use these new interface as they are not as intuitive as the traditional keyboard.

\subsection{Voice-based Text Entry}
Unlike the frustrating text dictation engines in the early days, modern deep learning-based language models, are capable of generating relatively accurate text from voice inputs. There are a number of popular models and toolkits can be integrated into AR systems, such as Whisper~\cite{whisperOpenAI}, Google Cloud Speech-to-Text, Amazon Transcribe, SpeechBrain, and etc.
For example, Zhang et al.~\cite{Zhang2022} were able to achieve an input speed of 23.5 wpm and a low error rate of 5.7\%, using the SWIFTER (Speech WIdget For Text EntRy) interface~\cite{Pick2016}.

Although voice-based text entry is intuitive and easy to use, it is not suitable for all situations. For example, it is not suitable for a noisy environment or when the user is in a public place. It is also inefficient to input the non-vocabularies, such as rare people's names, passwords, emojis, etc. Also, certain service providers process the users' voice data on their servers, which may raise privacy concerns apart from the common latency issue.

\subsection{Gaze-based Text Entry}
Many AR HMD-specific hands-free text entry methods have been proposed in recent years. Gaze-based text input allows users to input text by looking at the virtual keyboard and selecting the target character for a certain dwell time. For example, Mott et al.~\cite{Mott2017} used an eye-tracker to detect the user's gaze. This enabled the user to select the target character by looking at it for a cascading dwell time. This study achieved a typing speed of 12.39 wpm. The team of Lu et al.~\cite{Lu2020,LuDifengYu} also studied three gaze-based text entry methods: eye-blinks, dwell and swipe gestures. These three approaches achieved typing speeds of 11.95 wpm, 9.03 wpm, and 9.84 wpm respectively. 

\begin{figure}[h]
  \centering
  \includegraphics[width=0.9\linewidth, trim={5 10 0 0},clip]{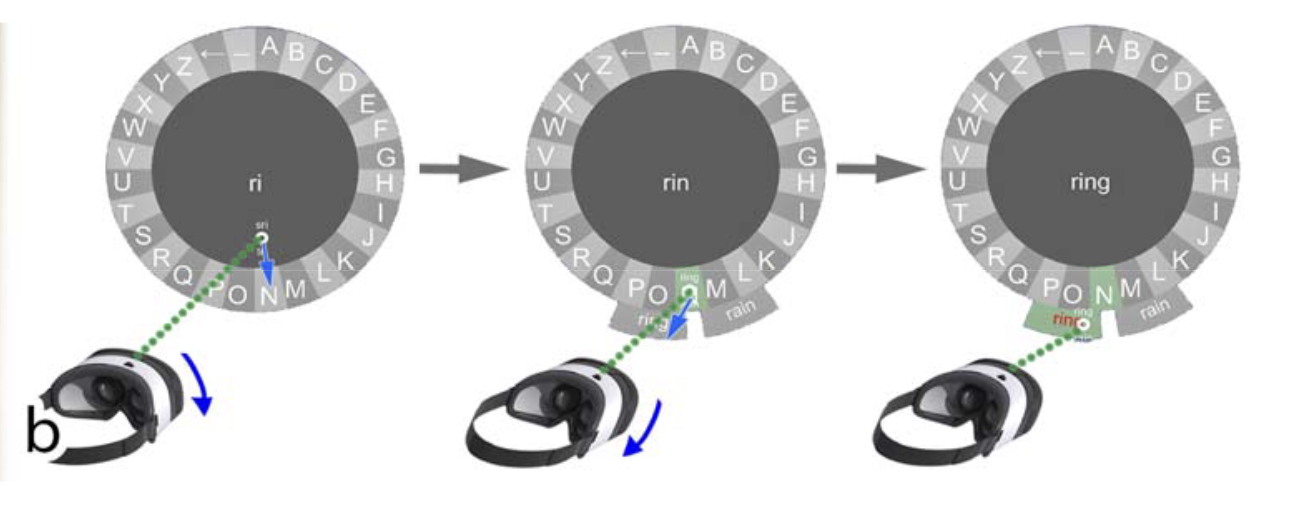}
  \caption{The interface of RingText~\citep{Xu2019}.}
  \label{fig:RingText}
\end{figure}

Incorporating eye-tracking and head-motion tracking, Xu et al.~\cite{Xu2019} proposed a gaze-based text entry method called RingText (see Figure~\ref{fig:RingText}). RingText is a circular-layout virtual keyboard that allows a user to select target characters by rotating the head to the target character. Users could reach a typing speed of 13.24 wpm with some training.

Gaze-based text entry requires a highly accurate eye-tracker, which is not always available in the current AR HMDs. The primary concerns are the accuracy eye-tracker. Taking Microsoft HoloLens 2 as an example, the built-in eye-tracker has a nominal spatial accuracy of 1.5 degree, which is fine for selecting larger holograms, but struggles to select smaller targets, such as keys on the virtual keyboard. Ergonomically, gaze-based text entry is more likely to cause eye fatigue.

\subsection{Mid-Air Tapping for Text Entry}
Mid-air typing is widely adopted in AR HMDs today. This is typically done by showing a floating holographic keyboard in front of the user, that the users could ``type'' or ``click'' as if they were ``typing'' a physical keyboard. 

A well known example is the MRTK (Mixed Reality Toolkits) keyboard. Markussen et al.~\cite{markussen2013selection} also evaluated three mid-air text input methods: hand-writing, typing on imaginary keyboards, and ray-casting on virtual keyboards. An OptiTrack\texttrademark system was used to track the user's hand movements. The best entry speed achieved from the study was only 13.2 wpm. Continuing this work, the authors then produced Vulture, a mid-air word-gesture keyboard~\cite{Markussen2014}. The user can input text by drawing a word in the air, achieving the best entry speed of 21 wpm.
Integrating auto-correcting to the mid-air text entry, Dudley et al.~\cite{Dudley2018} proposed the visualised input surface for augmented reality (VISAR) improved the typing speed using a single finger from 6 wpm to 18 wpm. 

As these holographic keyboards typically have the same layout to the physical ones, it requires minimal effort to get started with such interface. There are also drawbacks for these methods; apart from the low typing speed, these methods are unable to provide tactile feedback and can be prone to causing fatigue to the arms.

\subsection{Tap-on-Surface Text Entry}
In this paper, we refer tap-on-surface to those methods that allow users to gain tactile feedback, by projecting virtual keyboards on physical surfaces. Some wearable-based approaches (see Section~\ref{subsec:wearables}), such as PalmType~\cite{Wang2015}, QwertyRing~\cite{Gu2020} and TapType~\cite{Streli2022}, could also fall into this category. However, one noticeable downside with these methods is that they require users to wear additional devices, which are mostly custom made.

Recent works have shown that it is also possible to achieve similar output without any external hardware, as modern AR headsets are already packed with sensors. For example, MRTouch combined the depth and infrared camera in HoloLens 1 to perform real-time surface and hand detection, achieving an average position error of only 5.4mm~\cite{Xiao2018}. As the size of a key on a full-size physical keyboard is about 19mm, it is technically feasible to use MRTouch to perform tap-on-surface text entry, i.e., placing a keyboard on the table. 

\subsection{summary}
Based on previous research and our experience with a number of off-the-shelf AR headsets, the mid-air text entry method appears to be the most popular option. However, the lack of tactile feedback, low wpm, and ergonomically unfriendly interface, making it not quite the ideal solution for text entry in AR HMDs. On the other hand, the tap-on-surface text entry method is more promising as it is more intuitive and natural. However, it is still in its early stage and there are still many challenges to overcome.

However, we are yet to see any product that has successfully implemented this method. One of the main reasons is that the current AR HMDs have limited computing power, which makes it incapable of performing additional complex deep learning workflow (e.g., a deep Recurrent Network Network or a large Transformer model) other than those optimised OEM ones (e.g., hand gesture detection).

In this paper, we intend to exploit the data from on-board sensors, such as wide FOV (field of view) tracking cameras, colour cameras, depth cameras, and IMUs, which could be extracted and fed into dedicated neural networks to achieve real-time text entry that is solely based on the hand motions captured using the user perspective camera.

\section{Collection of AR Keystroke Detection Dataset} \label{sec:data_collection}
One big challenge is that there is simply no public dataset that is suitable for this task. There are some datasets that are related to hand motion/pose, such as the EgoGesture~\cite{EgoGesture} and NVGesture~\cite{NVgesture} datasets. However, gesture detection is quite different from keystroke detection and cannot be used in for our purpose. For example, gestures are often unique from each other (e.g., hand waving vs. thumb up), while keyboard typing motions are often similar to each other (e.g., pressing the key ``a'' vs. pressing the key ``s''). Also, they may not collected from the user's perspective, which is the most common use case for AR HMDs. As a result, we have to create our own dataset from scratch, which will be referred to as the AR Keystroke Detection Dataset (AKDD). As a starting point, this data will be limited to the English alphabet and the space key (27 keys in total).

The AR Keystroke Detection Dataset consists of ground truth record (csv files) and video sequences (mp4 files). The ground truth record contains the timestamp and the corresponding keystroke of each frame. The video sequences are recorded from the user's perspective, i.e., using a head-mounted camera or headset. The video sequences are recorded at 30 FPS with a resolution of $1920 \times 1080$. A python-based data logger was created, which monitors the keyboard input (ground truth) and aligns them with incoming video frames by timestamp.

The dataset is collected using the following protocol: The user is asked to sit in front of a table with a laptop placed on it. They are then instructed to wear a head-mounted camera, which has a wide field of view (FOV) and can capture both of their hands. Finally, the user is asked to type a list of 1,000 common English words (e.g., human, music and etc.) and 27 pangrams that are displayed on the laptop in a random order. The pangrams are the sentences that containing every letter of the alphabet at least once, such as \textit{`The quick brown fox jumps over the lazy dog”} and \textit{“pack my box with five dozen liquor jugs”}. The pangrams were pre-processed to convert all letters to lower case and replace all punctuations to spaces. This work has been approved by The Australian National University Human Research Ethics Committee with a protocol number of 2023/204. 

In total, we collected a total of 234,000 frames (130 mins). The distribution of samples for each class is shown in Table \ref{tab:sampleSize}. 

\begin{table}[hpt]
  \centering
  \caption{The sample size of each class (key) in the collected dataset.}
  \label{tab:sampleSize}
  \begin{tabular}{|c|c|c|c|c|c|}
    \hline
    0 - IDLE & 1 - "A"  & 2 - "B"  & 3 - "C"  & 4 - "D"  & 5 - "E"    \\
    183655   & 3140     & 1070     & 1297     & 1475     & 3839       \\
    \hline
    6 - "F"  & 7 - "G"  & 8 - "H"  & 9 - "I"  & 10 - "J" & 11 - "K"   \\
    1023     & 1276     & 1187     & 2391     & 954      & 994        \\
    \hline
    12 - "L" & 13 - "M" & 14 - "N" & 15 - "O" & 16 - "P" & 17 - "Q"   \\
    1952     & 1203     & 2074     & 2393     & 1526     & 1154       \\
    \hline
    18 - "R" & 19 - "S" & 20 - "T" & 21 - "U" & 22 - "V" & 23 - "W"   \\
    1845     & 1925     & 1531     & 1632     & 1069     & 1371       \\
    \hline
    24 - "X" & 25 - "Y" & 26 - "Z" & 27 - SPACE  & &  \\
    1474     & 1511     & 1575     & 7464       & & \\
    \hline
  \end{tabular}
\end{table}
The dataset used in this paper and the data logger will be made publicly available online. As the development of such dataset is still at its infancy, a continuously growing dataset will include other symbols and languages in the future. We believe the contribution from the community will play a vital role in the development of this dataset.

\section{Design of Real-Time Keystroke Identification Model} \label{sec:model_design}
\begin{figure*}[htp]
  \centering
  \includegraphics[width=.9\linewidth]{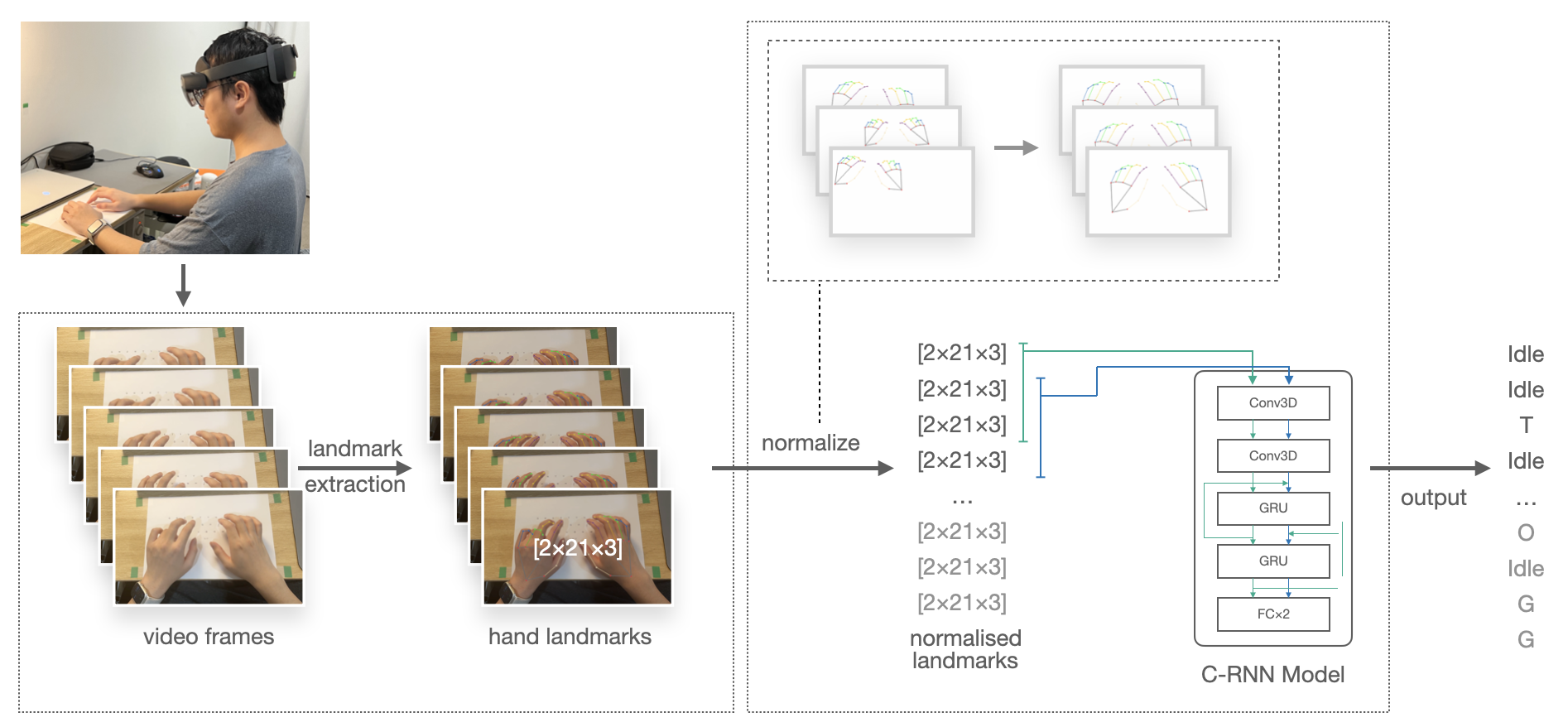}
  \caption{Overall architecture of the real-time keystroke identification model.}
  \label{fig:model_architecture}
\end{figure*}

This section presents a two-stage real-time keystroke identification model. The overall architecture of the model is shown in Figure~\ref{fig:model_architecture}. The first stage is hand landmark detection, which provides the world coordinates of hand landmarks. The second stage is keystroke detection and classification, which is used to detect the keystroke and classify it into one of the 27 keys or the ``idle'' state. Along with the model architecture, we also discuss the data augmentation techniques used in each stage, as well as the choice of training hyper-parameters, such as the loss function and the optimiser.

\subsection{Stage 1: Hand landmark Extraction}

\subsubsection{Raw frame augmentation and pre-processing}
A big challenge with AR text-inputting scenario is that people have different headsets and almost always wear them differently even if they are the same headset. This introduces a big issue that the camera is not always at the same position and angle relative to the keyboard, as seen in the limited training dataset. To increase model's ability to cope with different headsets and wearing styles, we applied a number of data augmentation techniques to the raw frames, including resizing, small-scale random cropping, rotation and affine transformations, which could simulate certain variations in hand postures.
\begin{figure}[h]
  \subfigure[Raw frame]{
    \begin{minipage}[c]{0.45\linewidth}
      \centering
      \includegraphics[width=\linewidth]{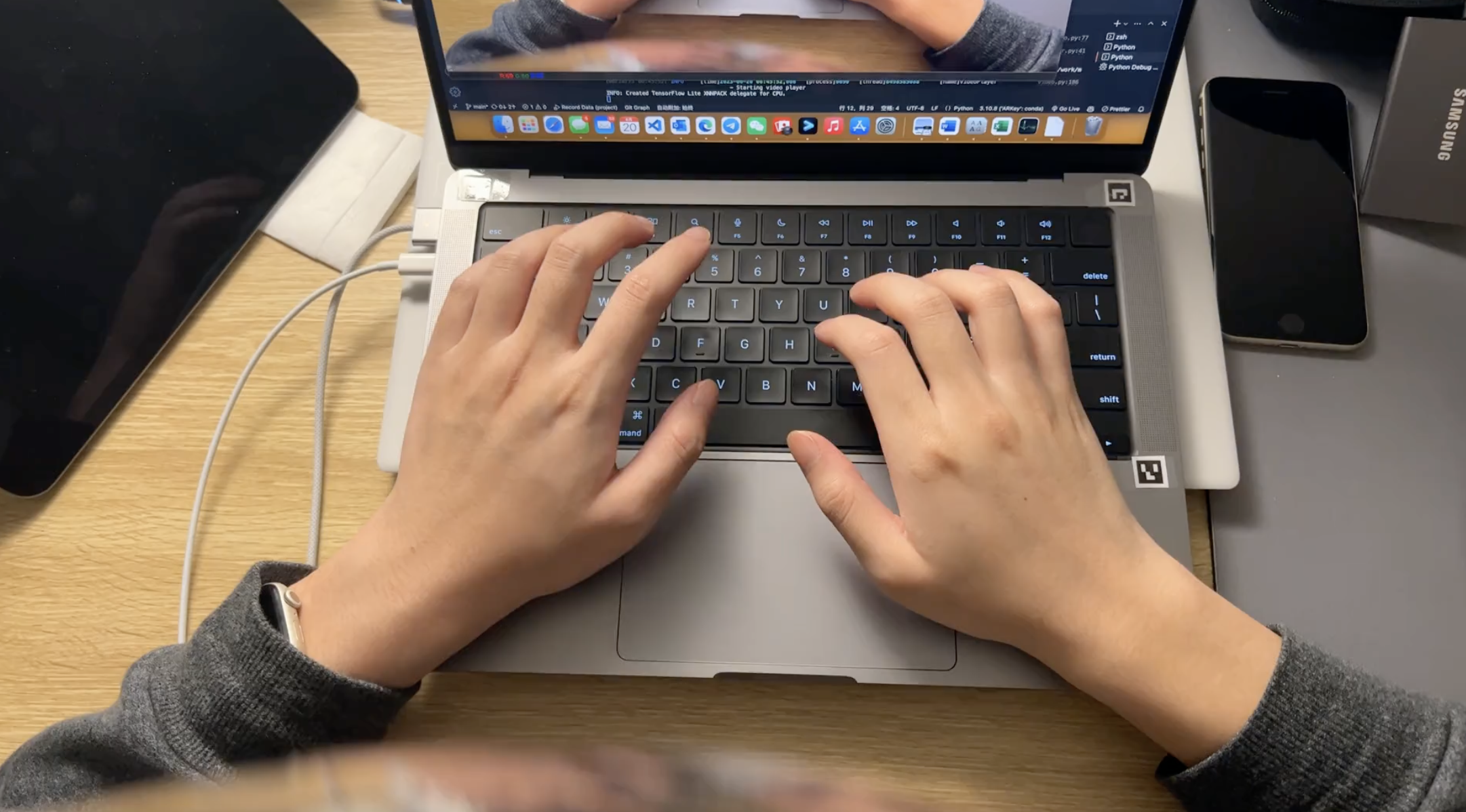}\vspace{4pt}
    \end{minipage}
  }
  \subfigure[Transformed frame]{
    \begin{minipage}[c]{0.45\linewidth}
      \centering
      \includegraphics[width=\linewidth]{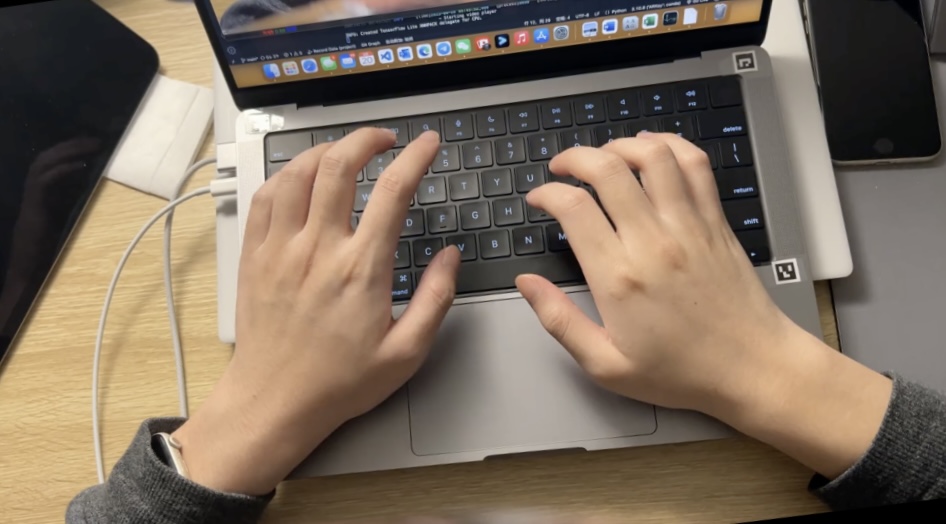}\vspace{4pt}
    \end{minipage}
  }
  \caption{An example of a transformed frame.
  \label{fig:frameCollected}}
\end{figure}

Apart from augmenting the raw frames, the ground truth labels also need to be transformed into a format that is suitable for the neural network. Important steps include one-hot encoding, class weight balancing, label smoothing, and sliding-window application.

\textbf{One-hot encoding:} This turns the N unique labels into N-dimensional vector. In this case, we have 28 unique labels, so each label was converted into a 28-dimensional vector, with the idle state being first, then alphabets and space following in alphabetical order. This also allows us  to smooth the label along the time axis.

\textbf{Class weight balancing:} Even though pangrams were used to maximise the balance between each letter of the alphabet, the distribution of these letters was far from even. In particular, the idle-state labels, representing moments where no key is pressed, significantly outnumbered the labels for the actual keystrokes. To address the imbalance in the distribution of the classes, we calculated class weights that we used later in the training process to adjust the loss function. The weight for each class is computed based on its representation in the dataset:
\[
  w_i = \frac{N}{k * n_i}
\]
where $w_i$ is the weight for class i, N is the total number of samples, k is the total number of classes, and $n_i$ is the number of samples in class i. This formula ensures that classes with fewer samples get higher weights, helping to counterbalance their under-representation in the training dataset. This is particularly important when training with a loss function, such as weighed cross entropy.

\textbf{Label Smoothing:} Since typing involves continuous finger movements to press a key rather than sudden presses, the states of about-to-press and just-pressed are highly similar and should have same label. Therefore, we applied a smoothing operation along the time axis to the labels. This operation is performed as follows:

Defining label $l_i$ as the one-hot encoded label for frame $i$ with its value being $y_{idle}$ for idle state or $y_{class_k}$ for the class $k$. We find all the $m$ and $n$ such that $\forall n \leqslant i \leqslant m, l_i = y_{class_k}$, $l_{m-1} = y_{idle}$, and $l_{n+1} = y_{idle}$. Then we apply a linear blend with size of $s$ to the label $ l_j \in [l_{m-s}, l_{m-s+1}, ..., l_{m-1}]$ and $l_k \in [l_{n+1}, ..., l_{n+s}]$:

\[
  l_j = l_j \cdot \frac{s - (m - j)}{s} + y_{class_k} \cdot \frac{m - j}{s}
\]
\[
  l_k = l_k \cdot \frac{s - (k - n)}{s} + y_{class_k} \cdot \frac{k - n}{s}
\]
This approach ensured that the labels correctly represented the gradual transition of the fingers pressing and releasing the keys.

\textbf{Sliding Window:} Sliding window is a common technique used when training spatial-temporal data to capture sequential dependencies and extract local patterns by splitting the continuous data into discrete chunks.
We used a window size of 128 with a step of 64 while preparing our dataset for training and validation.
\subsubsection{Hand Landmark Detection Models}
Hand landmark detection is a computer vision task that identifies and locates key landmarks on the hand, such as the fingertips and joints from images or videos. In the context of keystroke prediction, there are no dedicated hand tracking tools available. 

There are examples of using infrared cameras to track hand movements. For example,
Feit et al. used infrared tracking cameras with retro-reflective markers placed on the user's hand to analyse hand keystroke movements~\cite{Feit2016}. However, this approach is unsuitable for our study due to the additional hardware required. On the other side, there are also a few RGB camera-based hand landmarks tracking tools, such as OpenPose~\cite{openposehand}, 3DHandsForAll~\cite{lin2022ego2handspose}, and MediaPipe~\cite{mediapipe2020}.

\begin{figure}[h]
  {\includegraphics[width=.9\linewidth]{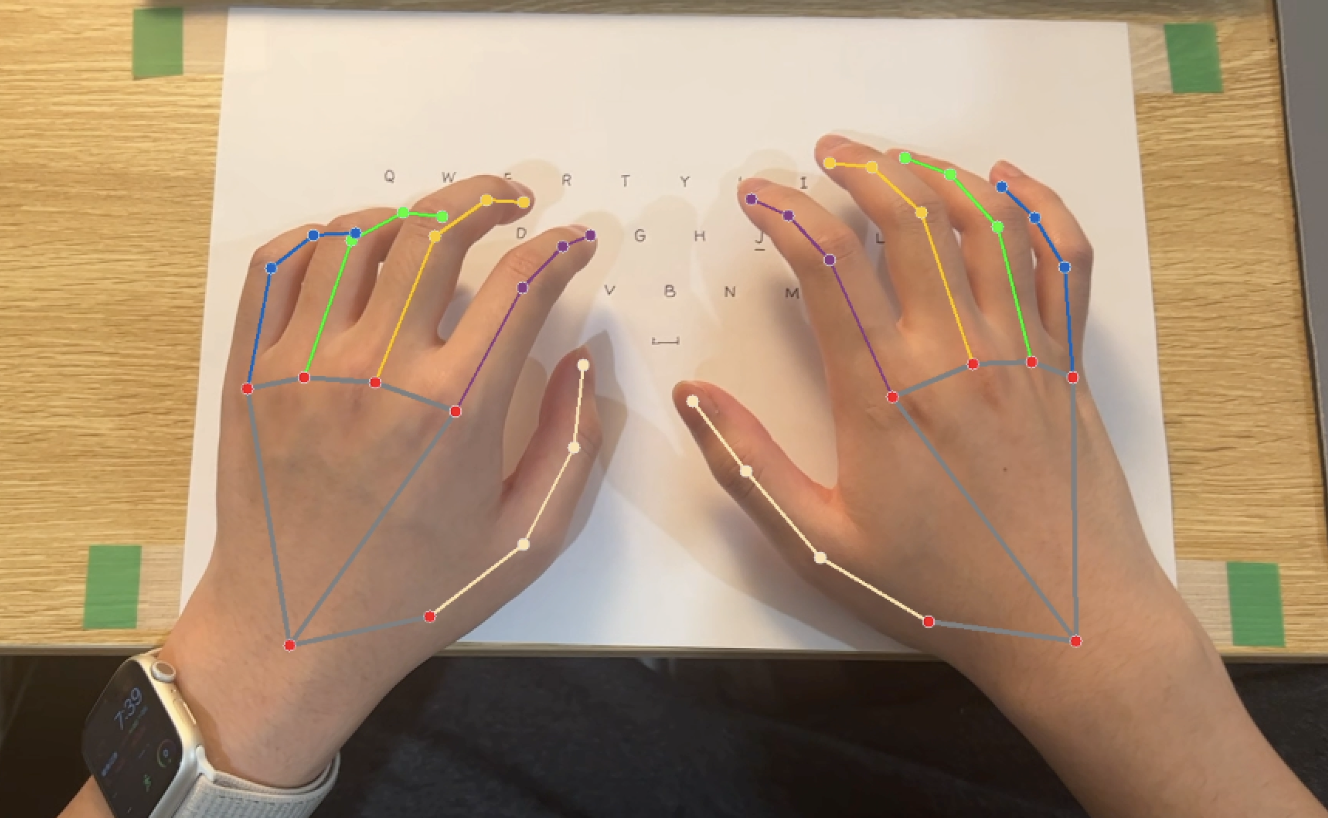}}
  \caption{Hand landmarks detected using MediaPipe.}
  \label{fig:mediapipe_example}
\end{figure}

In this early study, we used MediaPipe~\cite{mediapipe2020} as the hand landmark detection tool, as it is a more lightweight and efficient solution with easy-to-use APIs, making it an ideal choice for our task. MediaPipe can extract 21 hand keypoints in 3D coordinates from a single RGB image or video sequence, demonstrated in Figure~\ref{fig:mediapipe_example}. 


For each input frame, MediaPipe yields information on the number of hands in each frame, their respective handedness, and corresponding landmarks. A $(2 \times 21 \times 3)$ array was used to store the landmarks for each frame, assigning the first $(21 \times 3)$ for the left hand and the second for the right hand. In cases where only one hand was detected, we filled the unused array with zeros. Once all video frames were processed, we obtained a $(n \times 2 \times 21 \times 3)$ array for our data, where $n$ represents the number of processed consecutive frames. Apart from landmarks, we also had a corresponding $(n \times 28)$ array for our labels. 

There are two noticeable issues with MediaPipe predicted landmarks, which are referred to as ``world landmarks''. One issue is that, without stereo vision or reference markers, the depth information is not true depth, but relative depth. The other issue is that these ``world landmarks'', especially the fingertips, suffer from motion introduced by the typing activities and natural head movements. 
By default, MediaPipe normalises hand landmarks based on the wrist point of each hand individually, fails to adequately capture the relative positions between both hands. This limitation makes it challenging to accurately predict certain keys like R, T, F, G, V, B in the left hand zone, and U, Y, J, H, N, B in the right hand zone. This is because these predictions requires the understanding of the both hands' relative positions. 

In this work, we propose to normalise the method landmarks by scale and shift. Since the keypoint values extracted from MediaPipe are normalised within the range of $[0.0, 1.0]$ based on image width and height, variations in the distance between the camera and the hands can yield differing scales in landmark values. Moreover, hands may occupy different positions in the image, necessitating scale and shift transformations for data normalisation.
\begin{itemize}
    \item Shift: We establish a reference point, taken as the midpoint of the two wrist points. The coordinates of this reference point are subtracted from the coordinates of all other points, effectively shifting the position of the landmarks.
    \item Scale: Considering the relatively constant distance between the wrist and the root of the middle finger, we calculate this distance and then normalise the size of all keypoints by dividing their coordinates by this distance.
    \item Jitter removal: To eliminate the impact of jitters,
    we calculate the average coordinates of the reference point and the distance using a sliding window of 15 frames. Then, we normalise our data based on these averaged values.
\end{itemize}

At the end of stage-1, sequences of the ground truth labels and normalised hand landmarks are ready to be fed into the second stage of our workflow for keystroke detection and classification.

\subsection{Stage 2: Real-Time Keystroke Identification}
As task now becomes translating the sequence of hand landmarks generated from stage-1 into a sequence of keystrokes, a Sequence-to-Sequence (Seq2Seq) model becomes an obvious option. Seq2Seq model is a type of deep learning architecture that transforms an input sequence into an output sequence, capturing complex temporal dynamics and dependencies~\cite{sutskever2014sequence}. Several popular model families have been demonstrated to be effective in many sequence prediction tasks, such as speech recognition~\cite{chiu2018state} and video captioning~\cite{Aafaq2019}.

The Recurrent Neural Networks (RNN), including Long short-term memory (LSTM)~\cite{hochreiter1997long} and Gated Recurrent Unit (GRU)~\cite{cho2014learning} are powerful for sequence prediction tasks, such as speech recognition~\cite{chiu2018state} and video captioning~\cite{Aafaq2019}. This is because they can capture long-term dependencies; however, they typically suffer from vanishing/exploding gradients and are computationally expensive. 
Convolutional Neural Networks (CNN), such as Temporal Convolutional Network~\cite{bai2018empirical}, are another option. CNNs are good at identifying spatial hierarchies or local and global patterns within fixed-sized data, like images, but they don't inherently capture sequential dependencies in the data.
The attention-based Transformer~\cite{vaswani2017attention} also has been proven to be effective in many sequence prediction tasks.
However, transformers can be computationally expensive and memory intensive due to the self-attention mechanism's quadratic complexity with respect to input length.

It should be noted that the focus of this paper is verifying the feasibility of using deep learning to predict keystrokes from video data, rather than to fine-tune an optimal model. Consequently, we chose a model architecture that is simple to train, and lightweight enough to run in real time.

\begin{figure*}[h]
  {\includegraphics[width=.9\textwidth]{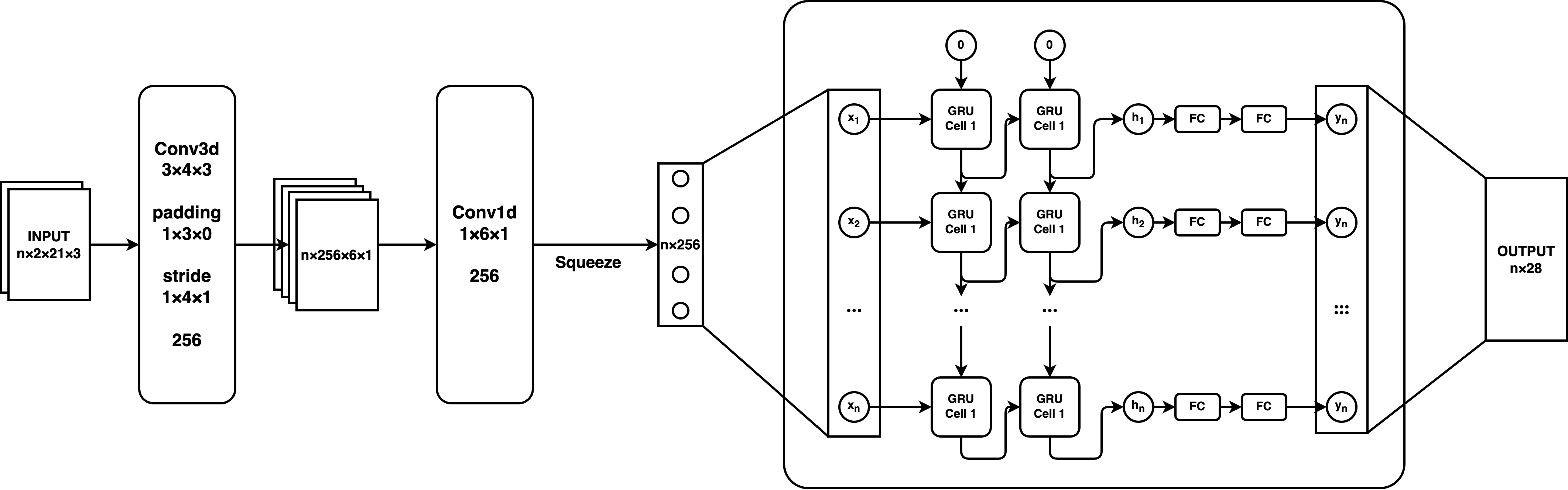}}
  \caption{The architecture of our model.}
  \label{fig:modelArchitecture}
\end{figure*}

We propose a Convolutional Recurrent Neural Network (C-RNN) model, which can capture local spatial features through CNNs and handle sequential data through RNNs. It is also computationally less expensive to train and deploy compared to transformers. Our C-RNN model has 2 $\times$ 3D convolutional layers for feature extraction and 2 $\times$ GRU layers to form a sequence-to-sequence architecture for our task (see Figure~\ref{fig:modelArchitecture}). There are also batch normalisation and dropout layers to prevent overfitting and accelerate training. Finally, the keystroke prediction will be made by 2 $\times$ fully connected layers with a softmax activation function.

This architecture allowed us to effectively extract spatial-temporal features from the input data and capture the temporal dependencies between different frames.

\subsubsection{Convolutional Layers}
We selected 3D convolutional layers to form the first two layers of our mode, which are adept at extracting spatial and temporal features from the input data, giving an abstract representation of the landmarks. We first reshape our data to $(b \times 2 \times n \times 21 \times 3)$, where $b$ is the batch size, $n$ is the window size, to separate the landmarks of the two hands, which are considered as two separate channels for the subsequent convolutional layers.

The reshaped data was then fed into a 3D Convolutional layer, with a kernel size of $(3,4,3)$, padding of $(1,3,0)$, and stride of $(1,4,1)$. The kernel size in the first dimension (that applies to the third dimension of data) is set to 3, so it can extract information from both the frame itself and its neighbouring frames. This temporal convolution allows us to capture the changes in hand movements over a short window. The second kernel size 4 with the stride in the third dimension groups the keypoints of each finger together, enabling the extraction of finger-specific features. And the last kernel size of 3 in the last dimension takes into account the x, y, z coordinates of each keypoint (landmark).

The output of the 3D Convolutional layer is then passed through another 3D Convolutional layer with a kernel size of $(1, 6, 1)$, which is equivalent to a 2D Convolutional layer with a kernel size of $(6, 1)$ in the last two dimensions. This layer takes the five fingers and the wrist as a whole, further extracting holistic hand features.

\subsubsection{Gated Recurrent Unit (GRU) Layers}
After two convolutional layers, the data is reshaped to $( b \times n \times f)$, where $f$ is the feature channel number that the second convolutional layer outputs. This reshaped data is then fed into two GRU layers forming a sequence-to-sequence architecture.

In our model, for sample $X_i$ in the batch, where $X_i = [x_1, x_2, \dots, x_n]$, for each $x_j$, we take itself and the hidden state $h^1_{j-1}$ from the frist GRU computed on the previous input $x_{j-1}$ as the input of the first GRU. Then we take the output $h^1_j$ and the hidden state $h^2_{j-1}$ from the second GRU computed on the previous input $h^1_{j-1}$ as the input of the second GRU. The output hidden state of the second GRU $h^2_j$ is the output of the whole GRU layer for $x_j$. The $h^2_j$ is then passed through two fully connected layers to produce the final output $y^{fc}_j$. Therefore, for sample $X_i$ in the batch, the output will be $Y_i = [y^{fc}_1, y^{fc}_2, \dots, y^{fc}_n]$.

This architecture allows for considering at least 3 frames for predicting a keystroke, having no limitation on the input window size, while taking into account the preceding data in the time series. The output of the GRU layer is then passed through two fully connected layers to produce the final output $y^{fc}_j$, i.e., the predicted keystroke for frame $x_j$.

\subsection{Loss Function and Optimiser} \label{subsec:model_training}

For the loss function, we tested both mean-squared error (MSE) and weighted cross-entropy (CE).
MSE was used as the loss function for results reported in this paper, as we found that the model trained with weighted cross-entropy loss function is more prone to predict unintended keystrokes, which is likely due to the fact that the cross-entropy loss function is more sensitive to the false negatives than the false positives.

As for the optimiser, we tested Adam/AdamW, SGD (Stochastic Gradient Descent), and Adagrad (Adaptive Gradient)~\cite{duchi2011adaptive}. We found that the model trained with the Adagrad optimiser with learning rate of 0.01 and incorporating the \textit{ReduceLROnPlateau} yielded better performance on the validation dataset.The \textit{ReduceLROnPlateau} is a learning rate scheduler, which dynamically adjusts the learning rate, reducing it by a factor of 0.5 whenever the validation loss does not decrease after a patience of 3 epochs. This allows the model to benefit from both coarse-grain and fine-grain optimisation stages, thereby providing better generalisation performance on the validation set.

\section{Results and Discussions} \label{sec:results}

\subsection{Training Results} \label{subsec:training_performance}
The train/test dataset was created using leave-n-recordings-out method, resulting 80/20 split. This process was repeated 5 times for 5-fold cross validation. We chose a label smoothing size of 3, a window size of 128 and step size of 64. We set an initial learning rate of 0.01, batch size of 64, and train the model for 100 epochs. The model converged at around 15 epochs and reached the lowest validation loss at around 25 epochs. 


We conducted 5-fold cross validation. For each fold, we choose the weight that achieved the lowest validation loss as the final model for subsequent benchmarks. All results reported here are averaged from 5-fold cross validation. our workflow also achieved a class-average accuracy of 89.18\% on the validation dataset, which indicates a promising feasibility in the context of a 28-class classification problem. The accuracy for each class is shown in Table \ref{tab:metrics}.

\begin{table}[ht]
\centering
\caption{Performance Metrics} \label{tab:metrics}
\begin{tabular}{lcccc}
\hline
\textbf{Class} & \textbf{Accuracy} & \textbf{Recall} & \textbf{Precision} & \textbf{F1-score} \\
\hline
IDLE & 96.09\% & 96.089\% & 95.014\% & 95.549\% \\
A & 91.48\% & 91.476\% & 91.791\% & 91.633\% \\
B & 90.06\% & 90.065\% & 92.873\% & 91.447\% \\
C & 85.98\% & 85.984\% & 95.789\% & 90.622\% \\
D & 87.17\% & 87.170\% & 88.984\% & 88.068\% \\
E & 90.21\% & 90.205\% & 94.914\% & 92.500\% \\
F & 82.55\% & 82.553\% & 91.080\% & 86.607\% \\
G & 88.50\% & 88.498\% & 95.026\% & 91.646\% \\
H & 83.70\% & 83.697\% & 95.833\% & 89.355\% \\
I & 94.52\% & 94.518\% & 89.917\% & 92.160\% \\
J & 89.27\% & 89.269\% & 93.541\% & 91.355\% \\
K & 91.59\% & 91.589\% & 80.493\% & 85.683\% \\
L & 90.75\% & 90.748\% & 85.449\% & 88.019\% \\
M & 87.12\% & 87.121\% & 89.009\% & 88.055\% \\
N & 92.23\% & 92.232\% & 90.453\% & 91.334\% \\
O & 84.05\% & 84.049\% & 91.106\% & 87.435\% \\
P & 95.14\% & 95.139\% & 87.540\% & 91.181\% \\
Q & 91.55\% & 91.549\% & 84.052\% & 87.640\% \\
R & 92.02\% & 92.025\% & 96.419\% & 94.170\% \\
S & 84.47\% & 84.472\% & 86.441\% & 85.445\% \\
T & 93.35\% & 93.347\% & 93.639\% & 93.493\% \\
U & 92.10\% & 92.099\% & 88.851\% & 90.446\% \\
V & 89.53\% & 89.535\% & 87.833\% & 88.676\% \\
W & 88.01\% & 88.008\% & 89.648\% & 88.821\% \\
X & 78.79\% & 78.788\% & 90.830\% & 84.381\% \\
Y & 93.46\% & 93.463\% & 82.861\% & 87.843\% \\
Z & 84.36\% & 84.363\% & 90.289\% & 87.226\% \\
SPACE & 89.05\% & 89.049\% & 93.700\% & 91.315\% \\
\hline
mean & 89.18\% &	89.18\%	&90.48\% &	89.72\% \\
\hline
\end{tabular}
\end{table}

Although the results still have room for improvement, they have demonstrated a strong potential to predict keystrokes from user perspective video data using deep learning techniques, such as the one proposed here. It is also evident that our workflow could benefit from a dataset with greater diversity.

A key attribute of our proposed model is its capability to effectively handle sequential data. The introduction of GRU layers in the current model significantly boosted the class average performance from 56\% to 89\%, highlighting the importance of temporal feature capture in our task.

\subsection{Inference Speed} \label{subsec:inference_speed}
The real-time functionality of our model hinges on the condition that processing a single frame, from capture to output, shouldn't exceed the time lapse between two consecutive frames. This ensures our system can cope with real-time video stream without inducing noticeable lag or delay. We benchmarked the average time taken to process a single $1920 \times 1080$ frame, from the extraction of hand landmarks to generating a keystroke prediction. The benchmark was performed in two setups, i.e., CPU-only and CPU + GPU. The CPU used here is an Intel i9-12900H, while the GPU is an RTX 2080Ti (laptop version).

When only using a CPU, the entire workflow achieved 32.2ms per frame or 31 FPS. Interestingly, the improvement when GPU was introduced was quite minimal, resulting in 31.1ms (~32 FPS). As our input video stream is 30 FPS (~33.33ms), the inference speed is sufficient for real-time text entry. It should be noted that the inference speed did not consider network latency. This is because the device was cable connected to the server machine via WLAN, which means the latency was negligible.

The breakdown of inference time suggested that MediaPipe needed approx. 28ms to process a frame. 
On the other hand, the fact that our C-RNN model only took ~4ms to generate a keystroke identification. This makes it possible to directly deploy the model on an AR headset, while swapping MediaPipe with the device's built-in hand tracking solution, such as the articulated hands on HoloLens 2.

\subsection{Normalised Levenshtein Distance} \label{subsec:nld}
In order to better understand the performance of our keystroke detection workflow in a real-world scenario, the Normalised Levenshtein Distance~\cite{Li2007nld} (NLD) was also applied to measure the similarity between the predicted text and the ground truth text.
The Levenshtein Distance is a measure of the similarity between two strings, taking into account the number of single-character edits (insertions, deletions, or substitutions) required to transform one string into the other~\cite{Levenshtein1966}. This is an intuitive way to understand the performance of our model in realistic typing scenarios.
Assuming there are the original text $T$ and the identified (predicted) text $I$, the Normalised Levenshtein Distance (the higher the better) can be calculated as:
\[
  \text{Normalised Levenshtein Distance} = 1 - \frac{\text{Levenshtein Distance}(T, I)}{\text{len}(T)}
\]

To exam the NLD, we intended to record typing sessions at various speed, i.e., 20/30/40/50 wpm. However, it was practically impossible for an average user to maintain a constant speed while typing naturally and reading text on the screen at the same time. Instead, we used the mean typing speed, calculated using a 10 second rolling average window. The real-time mean typing speed is shown on the display, so the user will adjust the typing speed. Post processing was also required to trim off the data if the typing speeds drops, for example, when a person occasionally stops to read the words, then the speed could drop down to 0 wpm. One of the authors conducted the benchmarking session, and four videos of typing on a keyboard at varying speeds of 20/30/40/50 wpm were recorded.

\begin{figure}[htp]
  \centering
  \includegraphics[width=0.92\linewidth]{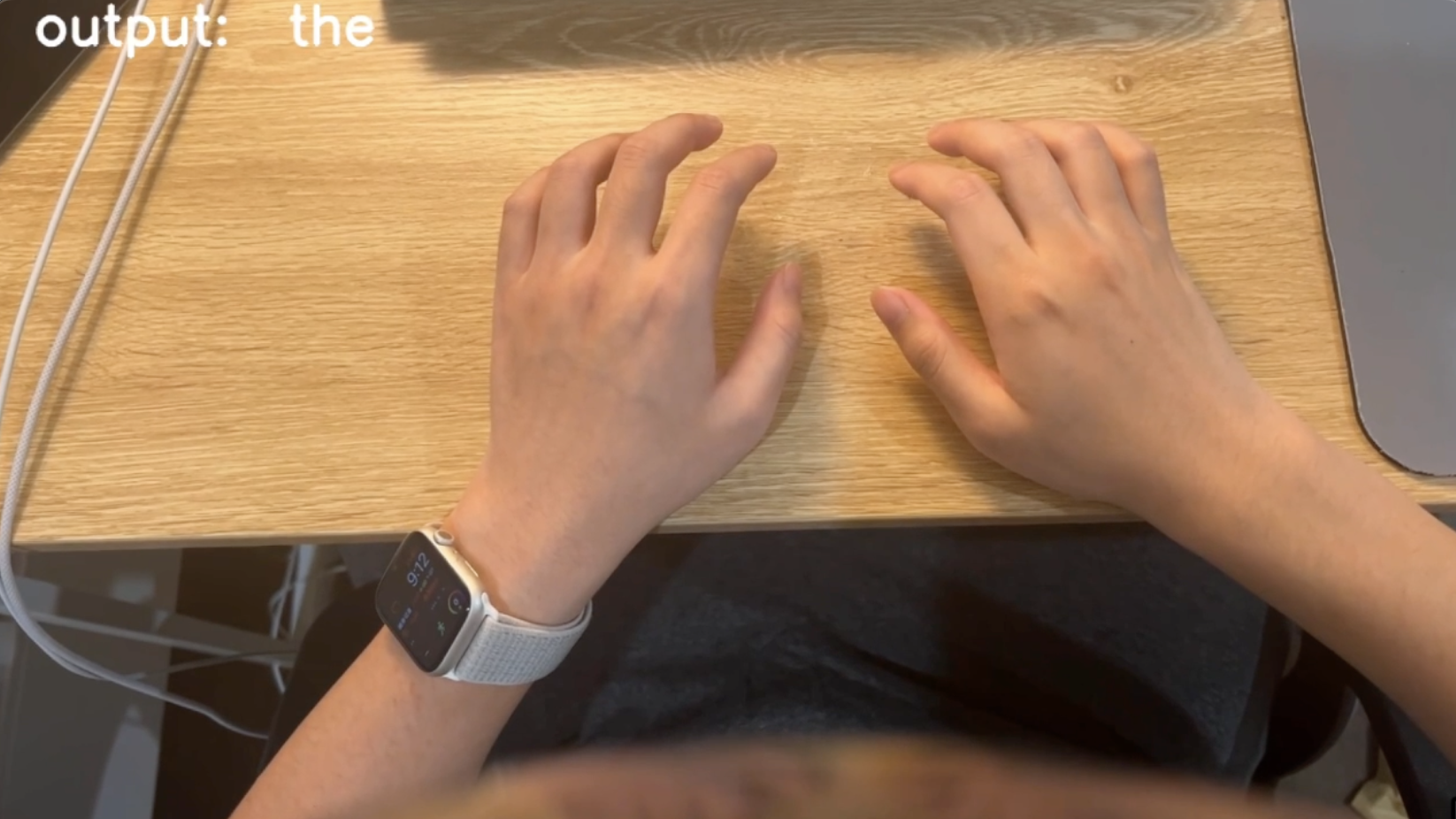}
  \caption{An example frame of the typing video.}
  \label{fig:typing_example}
\end{figure}
An example frame is shown in Figure~\ref{fig:typing_example}. The typing reference used were five pangram sentences. The results are shown in Table~\ref{tab:nld}
\begin{table}[h]
  \centering
  \caption{The Normalised Levenshtein Distance at different typing speeds.}
  \label{tab:nld}
  \begin{tabular}{|l|l|l|l|l|}
    \hline
    wpm           &  20      & 30      & 40      & 50\\ \hline
    NLD           &  96.22\% & 93.42\% & 91.05\% & 85.15\%\\ \hline
  \end{tabular}
\end{table}

The results suggested that our model performed consistently well before reaching a typing speed of 50 wpm (up to 96.22\% NLD)
The performance started to decline  passing the 50 wpm mark, which is likely due to the relatively small training dataset. This is expected to be greatly improved when the training dataset is expanded to include more variations.
It should be noted that 50 wpm is a reasonably fast typing speed even on a physical keyboard.

The examples below show the difference between our predicted text and the reference text (\textit{the quick brown fox jumps over the lazy dog}). The first line represents the reference text, and the second line illustrates the output from the model. The symbol ``␣'' denotes a space. The \textcolor{cyan}{text in cyan} represents the correctly identified keystrokes.

\begin{center}
  \texttt{\textcolor{cyan}{the␣quick␣b}r\textcolor{cyan}{o}w\textcolor{cyan}{n␣fox␣}\textcolor{cyan}{jum}p\textcolor{cyan}{s␣over}␣\textcolor{cyan}{the␣laz}y\textcolor{cyan}{␣dog}}

  \texttt{\textcolor{cyan}{the␣quick␣b}t\textcolor{cyan}{o}s\textcolor{cyan}{n␣fox␣}\textcolor{cyan}{jum s␣over the␣laz}u\textcolor{cyan}{␣dog}}
\end{center}


The example also shows that MSE-based model tends to produce unintended keys. We also trained the model using the weighted cross entropy loss function, which tends to miss keystrokes and performs worse. This is likely because the MSE loss function is more sensitive to the prediction error, which is more suitable for the keystroke detection task. It should be noted that fine-tuning these hyper-parameters is not the focus of this paper.

Overall, the results demonstrated that such type of workflow has a great potential to perform well in real-world scenarios. 

\section{Conclusions and Future Work} \label{sec:conclusions}


In conclusion, we've made the first attempt in conducting first-person perspective keystroke detection. We have created the first-of-its-kind dataset specifically tailored for training and evaluating such detection technology. Alongside this, we've established the first real-time deep learning workflow that facilitates real-time keystroke detection for AR applications, by combining hand landmark detection with a C-RNN architecture. Our early experiments showed promising results with an accuracy of up to 96.22\% at a lower speed (20 wpm) and 91.05\% at 40 wpm, i.e., the average typing speed when using a physical keyboard. Moreover, the inference speed of 32 FPS proves to be adequate for real-time text entry tasks. Although our main focus was on the AR domain, our model can be easily adapted to other applications, such as the removal of a physical keyboard when working on a tablet or a smartphone.

As we cast an eye towards future work, there are numerous promising paths of exploration. The first priority would be expanding the supported keys to include symbols and function keys, such as \textit{Shift} and \textit{Caps Lock}. Another priority is gathering more data to enrich our model's performance, as well as the exciting possibilities presented by experimenting with diverse data augmentation techniques. This could make the model significantly more flexible, such as being able to cope with different typing styles, compared to fixing index fingers to F and J keys in this work. With models trained with a much larger dataset, further user studies should consider a much larger and diverse participant pool to further validate and benchmark the model's performance, especially under uncontrolled real-world typing activities. As the current model only supports QWERTY keyboards, it may also worth considering how to support other types of keyboard layouts, such as AZERTY, Dvorak, etc. This could potentially be solved by enabling a ``keyboard layout'' calibration step into the model.

Another potential work is to use the built-in hand landmark detection features as stage-1 model, such as the articulated hands in HoloLens 2. This could have significant improvement due more accuracy hand landmarks at higher frequency, as AR headsets utilises high-frequency tracking cameras and dedicated signal processing unit to process them. This could eliminate the heavy computing cost from stage-1 and make it possible to deploy a more complex stage-2 model directly on the device. Additionally, we aim to conduct extensive hyper-parameter experimentation to optimise the model's performance. Finally, exploring complex and deeper network architectures could potentially open new doors for boosting the efficacy of our model.



\balance
\bibliographystyle{ACM-Reference-Format}
\bibliography{myref}

\end{document}